\documentclass{svproc}
\usepackage{url}
\usepackage{caption}
\usepackage{subcaption}
\usepackage{graphicx}
\usepackage{tabularx}

\begin{document}
\mainmatter              
\title{Real-Time Human Fall Detection using a Lightweight Pose Estimation Technique}
\titlerunning{RTHFDULPET}  
%
\author{Ekram Alam\inst{1,3} \and Abu Sufian\inst{2} \and
Paramartha Dutta \inst{3} \and Marco Leo \inst{4} }
\authorrunning{Ekram Alam et al.} 
%
%
\institute{Department of Computer Science, Gour Mahavidyalaya,West Bengal, India 
\and
Department of Computer Science, University of Gour Banga, India 
\and Department of Computer and System Sciences, Visva-Bharati University,  India
\and National Research Council of Italy, Institute of Applied Sciences and Intelligent Systems, 73100
Lecce, Italy }

\maketitle              

\begin{abstract}
The elderly population is increasing rapidly around the world. There are no enough caretakers for them. Use of AI-based in-home medical care systems is gaining momentum due to this. Human fall detection is one of the most important tasks of medical care system for the aged people. Human fall is a common problem among elderly people. Detection of a fall and providing medical help as early as possible is very important to reduce any further complexity. The chances of death and other medical complications can be reduced by detecting and providing medical help as early as possible after the fall. There are many state-of-the-art fall detection techniques available these days, but the majority of them need very high computing power. In this paper, we proposed a lightweight and fast human fall detection system using pose estimation. We used `Movenet' for human joins key-points extraction. Our proposed method can work in real-time on any low-computing device with any basic camera. All computation can be processed locally, so there is no problem of privacy of the subject. We used two datasets `GMDCSA' and `URFD' for the experiment. We got the sensitivity value of 0.9375 and 0.9167 for the dataset `GMDCSA' and `URFD' respectively. The source code and the dataset GMDCSA of our work are available online to access.
\keywords{Fall Detection, Pose Estimation, GMDCSA, Movenet, Lightweight Fall Detection, Real-time Fall Detection}
\end{abstract}
\section{Introduction}
Human fall is one of the major reasons for hospitalization in elder people around the world \cite{alam2022vision}. Detection of human falls is very vital so that medical help can be provided as early as possible. Human fall detection can be done using wearable, ambient, or vision sensors \cite{wang2020possible}. Vision-based fall detection system is more suitable, especially for elder people \cite{gutierrez2021comprehensive}. There is no need to attach the vision sensor to the body like wearable sensors. Wearable sensors need to be charged frequently whereas vision sensors can work on a direct home power supply. Human fall detection is one of the useful application of computer vision \cite{alam2021leveraging}, \cite{wang2020elderly}. In this paper, we have proposed a lightweight human fall detection system using pose estimation \cite{munea2020progress}, \cite{chen2020monocular}. We have used a lightweight and fast pose estimation model `Movenet Thunder' \cite{bajpai2021movenet} for our work. The main contributions of this work are as given below.

\begin{description}
	
\item [Real Time] `Movenet' processes the video with 30+ FPS \cite{tensorflowMoveNetUltra} (real-time) in the majority of current low computing devices like mobile phones, laptops, and desktops. So the proposed system can work in real-time on these devices. We tested our work on an average computing laptop with inbuilt webcam. 

\item [Lightweight] The proposed system does not required very high computing power and can work on any normal laptop/desktop or mobile device.
\item [Local Computation] All computation can be processed locally. There is no personal data (images/frames) transfer from edge \cite{sufian2021deep} to the cloud and vice versa. Only the output (fall) is sent to the caretaker center for necessary medical help. In this way, our system also preserves the privacy of the subject. 
\item [GMDCSA Dataset] A new fall detection dataset named GMDCSA was introduced. 
	
\end{description}

The rest of the paper is structured as follows. Section \ref{S_RelatedWork} describes related work briefly. Section \ref{S_Background} gives an overview of the pose estimation using `Movenet'. Section \ref{S_Methodology} discusses the methodology of our work. Section \ref{S_Dataset} describes the datasets which were used in this work. Section \ref{S_Result} provides the results of this work in the form of different metrics. Finally, section \ref{S_Conclusion} concludes the proposed work with possible future scopes.

\section{Background Study}
\label{S_Background}
\begin{figure}[!htb]
	
	\centering
	
	\includegraphics[scale=0.40]{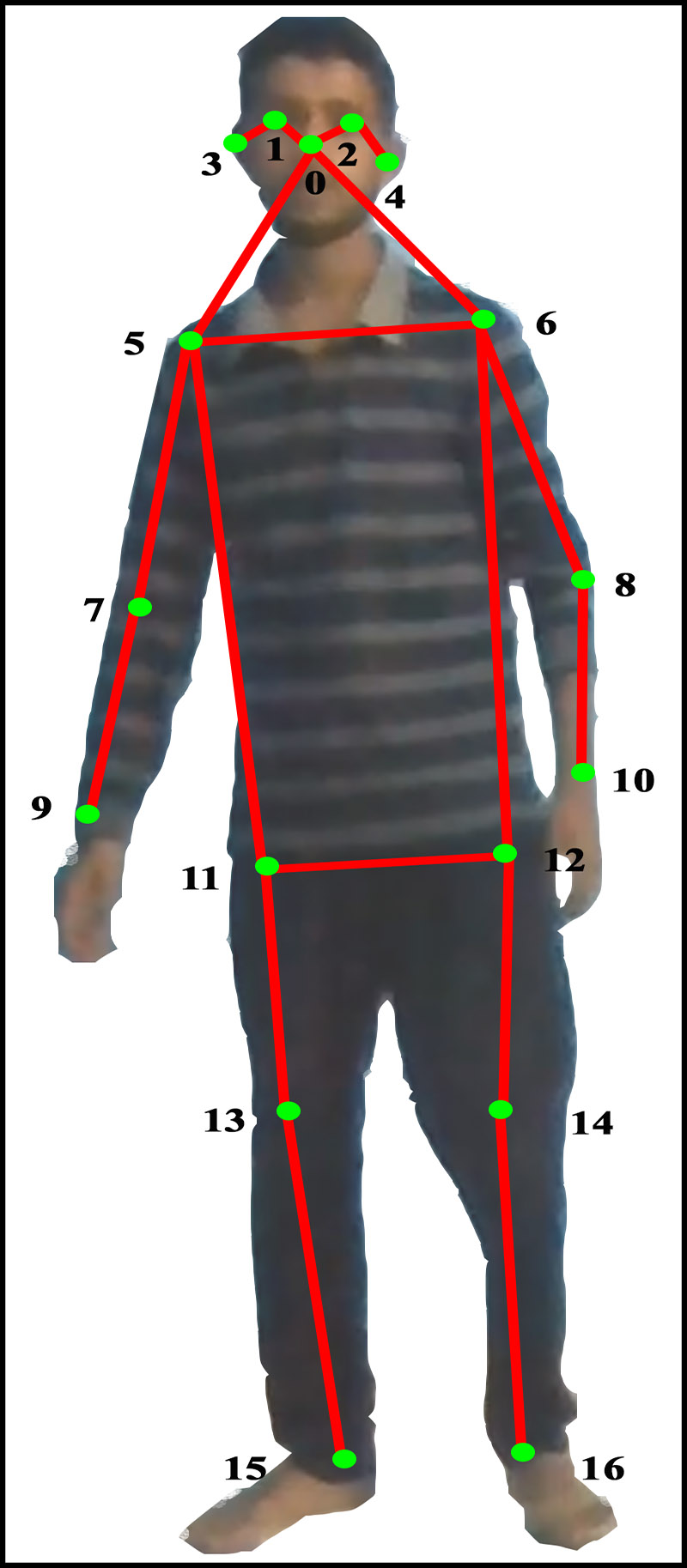}
	\caption{17 Keypoints of the Movenet pose estimation model}
	
	\label{F_pose}
	
\end{figure}

We have used a lightweight pose estimation model named `Movenet Thunder' \cite{bajpai2021movenet}.  This model accepts an RGB frame or image of the size 256 x 256  and extracts the normalized coordinate and confidence values of the 17 key-points of the human body joints. The 17 key-points are shown in Figure \ref{F_pose}.  
The indices (from 0 to 16), Keypoints, Y values, X values, and confidences value of a sample image are shown in Table. \ref{T_Index}. 
\begin{table}[!htb] 
	\centering
	\small
	\caption{Index, Keypoint, Y value, X value and confidence of an sample image}
	\label{T_Index}
	\begin{tabular}{|p{1.3cm}p{2.5cm}p{2cm}p{2cm}l|} \hline
		\textbf{Index}& \textbf{Keypoint}	& \textbf{Y Value}& \textbf{X Value} & \textbf{Confidence}\\ \hline
		0	&	Nose	&		0.22416662 &		0.579579   	&	0.7201656 \\
		1	&	Left Eye&		0.20926172 	&	0.5974146  &		0.8043867 \\
		2	&	Right Eye 	&	0.20485064 	&	0.5642889  	&	0.5905826 \\
		3	&	Left Ear	&	0.22323    	&	0.6126661  	&	0.7964257 \\
		4	&	Right Ear	&	0.21771489 	&	0.5370738  		&	0.7529471 \\
		5		&	Left Shoulder 	&	0.3235461 &0.6375601 &0.8950565 \\
		6		&	Right Shoulder	&	0.2964768 &0.48282918&0.65825576\\
		7		&	Left Elbow		&	0.43468294&0.63684213&0.7667525 \\
		8		&	Right Elbow		&	0.42770475&0.4406372 &0.8829603 \\
		9		&	Left Wrist		&	0.54110587&0.6462866 &0.6282949 \\
		10		&	Right Wrist		&	0.5392799 &0.42464092&0.8215329 \\
		11		&	Left Hip		&	0.54277164&0.57565194&0.85804665 \\
		12		&	Right Hip		&	0.53679305&0.48321638&0.88962007 \\
		13		&	Left Knee		&	0.69595444&0.609515  &0.8796475 \\
		14		&	Right Knee		&	0.7019378 &0.46842176&0.6786141 \\
		15		&	Left Ankle		&	0.85588527&0.56420994&0.7951814 \\
		16		&	Right Ankle		&	0.8588409 &0.47616798&0.82729894\\ \hline

	\end{tabular}
	
\end{table}	
The values of y,x, and confidence are normalized from 0 to 1. The top left position is the origin(0,0) and the bottom right position has the value (1,1). When the keypoints are clearly visible then confidence tends to 1 (100\%) otherwise it tends to 0 (0\%).

\section{Related Work}
\label{S_RelatedWork}

This section briefly describes some recent related works. Asif et al. \cite{asif2020sshfd} introduced a single-shot fall detection technique using 3D poseNet. Chen et al. \cite{chen2022video} proposed a 3D posed estimator which was used as input for the fall detection network. Apicella and Snidaro \cite{apicella2021deep} proposed a fall detection method based on CNN, RNN and PoseNet pose estimation. Leite et al. \cite{leite2021three} introduced a multi (three) channel CNN-based fall detection system. Optical flow, pose estimation, and visual rhythm were used as inputs for three different streams of the CNN. OpenPose \cite{cao2017realtime} was used for pose estimation. Chen et al. \cite{chen2022elderly} proposed a fall detection system using the Yolov5 network \cite{yolov5}. Liu et al. \cite{liu2022fall} proposed a fall detection system based on BlazePose-LSTM. This system was introduced especially for seafarers. Beddiar et al. \cite{beddiar2022fall} introduced a work based on the angle formed by the line from the centroid of the human face to the centroid of the hip to the line formed from the centroid of the hip to the horizontal axis. Amsaprabhaa et al. \cite{amsaprabhaa2023multimodal} proposed a multimodal gate feature-based fall detection system.

\section{Methodology}
\label{S_Methodology}
\begin{figure}[!htb]
	
	\centering
	
	\includegraphics[scale=0.55]{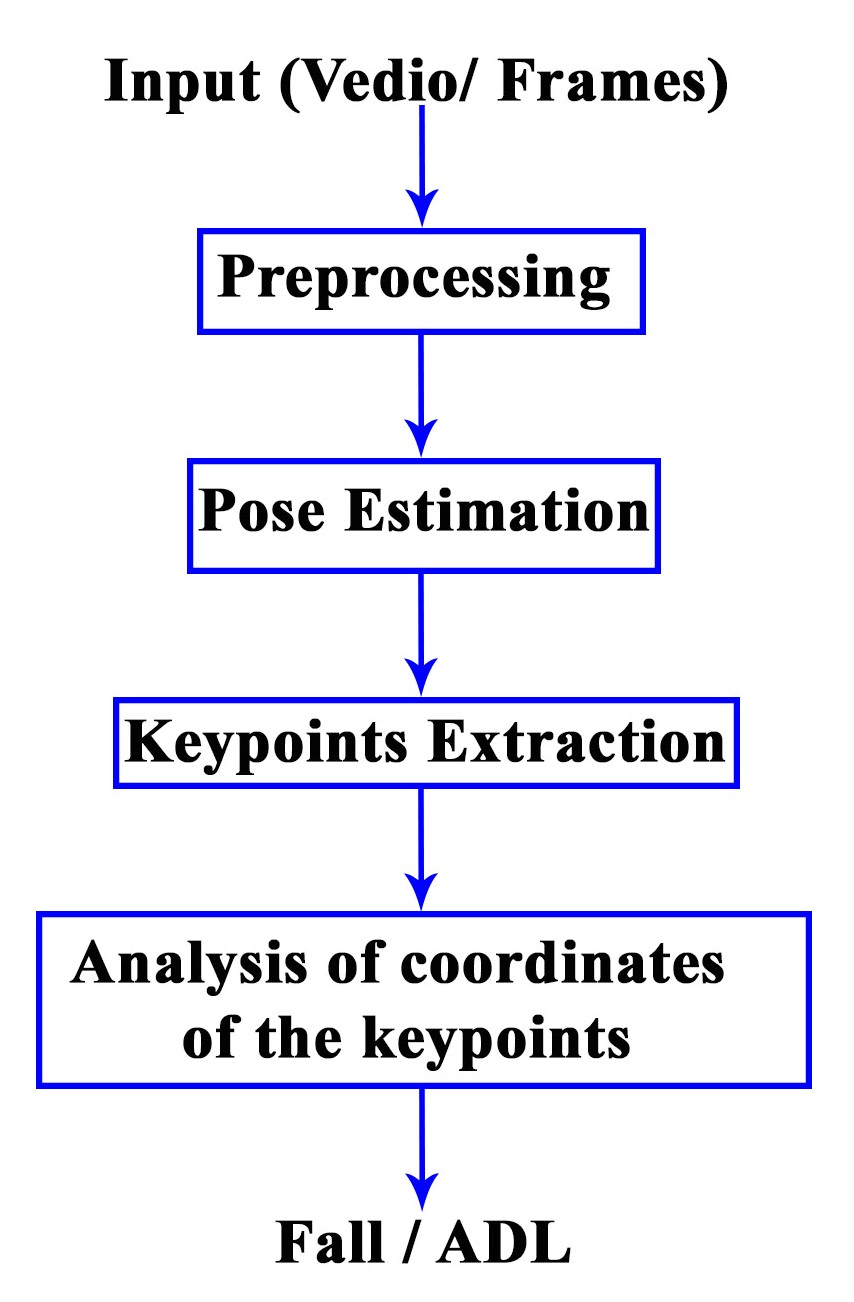}
	\vspace{-5pt}
	\caption{Proposed work methodology}
	
	\label{F_Methodology}
	
\end{figure}

The methodology of the proposed work is shown in Figure \ref{F_Methodology}.
\vspace{-5pt}
The input can be an image, frames of a video, or the live video stream. The input images / frames were resized to 256 x 256 as preprocessing before feeding it  to the Movenet. After preprocessing, pose estimation was done using the Movenet. The Movenet extracts the key-point co-ordinates with their confidence score as shown in Table \ref{T_Index}. Confidence score can vary from 0 (0\%) to 1 (100\%) for each keypoints. If all key-points with very low confidence are also used for fall detection then it might select the wrong keypoints and this will reduce the performance of the system. If the high confidence value threshold are used it may ignore some good keypoints  which might be useful for the detection of the fall. After experimenting with different threshold values of the confidence score, we finally selected 0.5 as threshold value because it gave good results. We have selected only those key-points whose confidence scores are greater than 0.5. The fall activity and sleeping activity are very similar and there are high chances of detecting a sleeping activity as fall. If there is a sleeping like activity on the floor then the system should detect it as fall activity, but if there is a sleeping like activity on the bed then the system should detect it as ADL (not fall) activity. To filter out this we compared the approximate y value of the top of the bed  with the y value of the nose, eyes, ears, shoulders, elbows and wrists (upper body part). If the y value of these key-points (upper body part) are greater than the approximate y value of the top of the bed, then the activities of these frames are not fall and filtered out for the fall detection.
\begin{figure}[!htb]
	
	\centering
	
	\includegraphics[scale=0.75]{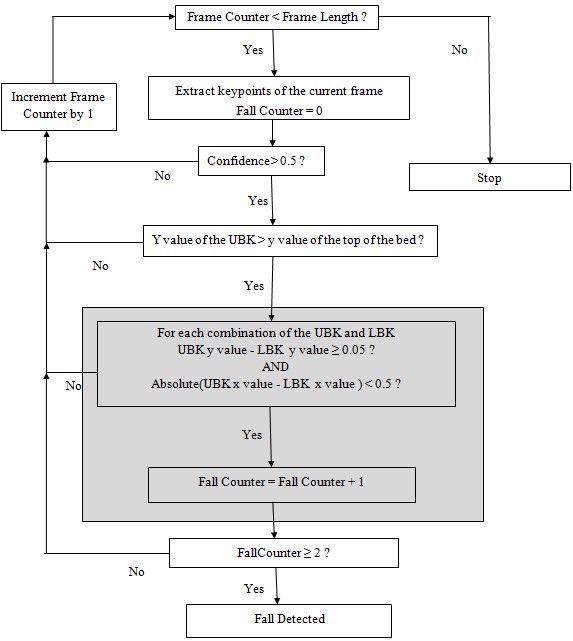}
	\caption{Proposed work methodology flowchart}
	
	\label{F_Analysis}
	
\end{figure}
After that the coordinates (x, y values) of the keypoints of the upper body parts were compared with the coordinates  of key-points of the lower body parts(hips, knees, ankles). If the differences of the y value of the upper body keypoints (UBK) with lower body keypoints (LBK) is less than or equal to 0.05 (threshold-y) and the absolute differences of the x value is greater than 0.5 (threshold-x) then there is a chance of a Fall and the fall counter is increased by one. The selection of the values for the threshold-y and threshold-x were done after doing many experiments with different values. These values gave the best results. If in the next frame this is false then counter reset to 0. If this happens continuously for 2  or more frames (minimum counter value 2), then the system detects it as a fall, and a fall alert is sent. Detail of the analysis is shown in Figure \ref{F_Analysis}. The source code of the proposed work is available here \url{https://github.com/ekramalam/RTHFD}.

\section{Dataset}
\label{S_Dataset}
We used two datasets for the proposed experiment, the URFD \cite{kwolek2014human} dataset and a dataset (GMDCSA) created by us. The URFD dataset contains 40 ADL (not fall) activities and 30 fall activities. The GMDCSA dataset contains 16 ADL (not fall) activities and 16 Fall activities. The GMDCSA dataset has been created by performing the fall and the ADL activities by a single subject wearing different set of clothes. The web camera of a laptop (HP 348 G5 Laptop : Core i5 8th Gen/8 GB/512 GB SSD/Windows 10) was used to capture the activities. The description of the ADL and Fall video sequences of the GMDCSA dataset are shown in Table \ref{T_ADLact} and \ref{T_Fallact} respectively. The link to access this dataset is as follows \url{https://drive.google.com/drive/folders/1ohDEXki8Wz12cJ1XzyKIK4T6y1_hAf3p?usp=sharing}. 

\begin{table}[h] 
	\scriptsize
	\caption{GMDCSA Dataset: ADL Activities}
	\label{T_ADLact}
	\begin{tabularx}{\linewidth}{|p{1cm}|p{1cm}|X|} \hline
		
		\textbf{File Name} &\textbf{Length} &\textbf{Description}  \\ \hline
		01.mp4 &08 sec &Sitting on the bed to sleeping right side on the bed. Face towards camera  \\ 
		02.mp4 &06 sec &Sitting on the bed to sleeping left side on the bed. Face towards camera.   \\ 
		03.mp4 &06 sec &Sitting on the bed to sleeping left side on the bed. Face towards ceiling.  \\ 
		04.mp4 &05 sec &From sleeping left side on the bed (Face towards ceiling) to sitting on the bed.  \\ 
		05.mp4 &10 sec &Coming to the bed and reading book in sitting position.  Front view of the subject. One leg folded.\\ 
		06.mp4 &12 sec &Sitting on the bed (front view) to reading the book while supporting towards the wall. Side view, Leg straight. \\ 
		07.mp4 &07 sec &Reading book while sitting on the chair. Front view.  \\ 
		08.mp4 &06 sec &Walking in the room.  \\ 
		09.mp4 &04 sec &Reading book while walking in the room.  \\ 
		10.mp4 &03 sec & Reading book while walking in the room.  \\ 
		11.mp4 &09 sec & Walking to sitting on the  chair and then reading book.  \\ 
		12.mp4 &07 sec & Reading book while sitting on the chair to stand up and keeping the book on the chair and going out of the room. \\ 
		13.mp4 &06 sec & Walking to sitting on the chair (side view).  \\ 
		14.mp4 & 05 sec & Sitting on the chair (side view) to walking.    \\ 
		15.mp4 & 07 sec & Walking to picking a mobile phone from the ground  and then sitting on the chair. \\ 
		16.mp4 & 03 sec & Picking the mobile phone from the ground while sitting on the chair  \\ \hline

	\end{tabularx}
\end{table}

\begin{table}[h] 
	\scriptsize
	\caption{GMDCSA Dataset: Fall Activities}
	\label{T_Fallact}
	\begin{tabularx}{\linewidth}{|p{1cm}|p{1cm}|X|} \hline
		
		\textbf{File Name} &\textbf{Length} &\textbf{Description}  \\ \hline
		01.mp4 &06 sec &Falling from sitting on the chair to the ground. Left side Fall. Full body not visible. \\ 
		02.mp4 &06 sec & Falling from sitting on the chair to the ground. Left side Fall. Full body not visible.  \\ 
		03.mp4 &05 sec & Falling from sitting on the chair to the ground.  Left side fall. \\ 
		04.mp4 &04 sec & Falling from sitting on the chair to the ground.  Right side fall.  \\ 
		05.mp4 &05 sec &Walking to falling. Right side fall.\\ 
		06.mp4 &05 sec & Walking to falling. Right side fall. \\ 
		07.mp4 &05 sec & Walking to falling. Right side fall.  \\ 
		08.mp4 &04 sec & Walking to falling. Left side fall.  \\ 
		09.mp4 & 04 sec & Standing position to falling. Forward Fall. Full body (head) not visible. \\ 
		10.mp4 & 04 sec & Standing position to falling. Forward Fall. Full body (right eye) not visible.   \\ 
		11.mp4 & 06 sec & Standing position to falling. Backward Fall. \\ 
		12.mp4 & 06 sec & Standing position to falling. Backward Fall.  \\ 
		13.mp4 & 04 sec & Standing position to falling. Backward fall. Full body (head ) is not visible. \\ 
		14.mp4 & 05 sec & Standing position to falling. Backward fall. Full body (both ankles) is not visible.  \\ 
		15.mp4 & 06 sec & Sitting on the chair (side view) to right side fall.  \\ 
		16.mp4 & 06 sec &  Sitting on the chair (side view) to left side fall.  \\ \hline

	\end{tabularx}
\end{table} 

%
%
%
%
%

\section{Result}
\label{S_Result}
\begin{figure}
	\centering
	
	\begin{subfigure}{.5\textwidth}
		\centering
		\includegraphics[width=.75\linewidth]{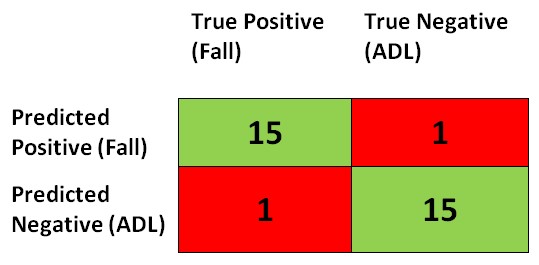}
		\caption{GMDCSA}
		
	\end{subfigure}%
	\hfill
	\begin{subfigure}{.5\textwidth}
		\centering
		\includegraphics[width=.75\linewidth]{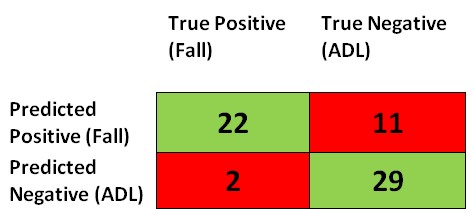}
		\caption{URFD}
		
	\end{subfigure}
		\vspace{-5pt}
	\caption{Confusion Matrix}
	\label{F_ConfMx}
		\vspace{-7pt}
\end{figure}

The performance of a model can be measured using different evaluation metrics \cite{alam2022vision} like sensitivity, specificity, precision, etc. The values of True Positive (TP), True Negative (TN), False Positive (FP), and False Negative (FN) are needed to calculate the values of these metrics. The value of TP, TN, FP, and FN can easily be found from the confusion matrix as shown in Figure \ref{F_ConfMx}. 
The value of TP, TN, FP, and FN for the two dataset GMDCSA and URFD are shown in Table \ref{T_TPtoFN}.
\begin{table}[h!]
	\centering
	\vspace{-5pt}
	\small
	\caption{The value of TP, TN, FP, and FN for the dataset GMDCSA and URFD}
	\label{T_TPtoFN}
	\begin{tabular}{|l|l|l|l|l|} \hline
		\textbf{Dataset}&\textbf{TP}&\textbf{TN} &\textbf{FP}&\textbf{FN}\\ \hline
		
 GMDCSA&15 & 15& 1&1 \\
 URFD  & 22&29 & 11&2  \\ \hline
	\end{tabular}
	\vspace{-5pt}	
\end{table}
Table \ref{T_Results} shows the values of sensitivity, specificity, precision, false positive rate, false negative rate, accuracy, and F1 Score.
\begin{table}[h!]
	\centering
	\vspace{-5pt}
	\small
	\caption{Results of the experiment}
	\label{T_Results}
	\begin{tabular}{|l|l|l|l|} \hline
		\textbf{Metric}&\textbf{Expression} &\multicolumn{2}{c|}{Dataset}\\ \cline{3-4}
		&&\textbf{GMDCSA}&\textbf{URFD}\\ \hline
		Sensitivity &	$$TP / (TP + FN)$$ &	0.9375 &0.9167\\
		Specificity	&	TN / (FP + TN) & 0.9375 &0.7250\\
		Precision &	TP / (TP + FP) &	0.9375 &0.6667\\
		False Positive Rate &	FP / (FP + TN)&	0.0625 &0.2750\\
		False Negative Rate	 &	FN / (FN + TP)& 0.0625 &0.0833\\
		Accuracy &	(TP + TN) / (P + N)&	0.9375 &0.7969\\
		F1 Score &	2TP / (2TP + FP + FN)&	0.9375 &0.6216\\ \hline

	\end{tabular}
\end{table}
 These values can be calculated using the values of TP, TN, FP, and FN as shown in Table \ref{T_TPtoFN}. Table \ref{T_Results} also shows the expressions of the corresponding metrics. The Sensitivity is more important than other metrics for any medical classification problem like human fall detection. The values of sensitivity are 0.9375, and 0.9167 for GMDCSA and URFD respectively. These values are good enough for a lightweight system. The specificity for GMDCSA is 0.9375 whereas for URFD it is 0.7250. The performance of our model is better using the GMDCSA dataset than the URFD dataset. This may be because the ADL activities of URFD contained many complex falls-like activities. Some of these activities were classified wrongly as falls by our system. 
 \begin{figure}[!htb]
 	\centering 
 	\begin{subfigure}{0.3\textwidth}
 		\includegraphics[width=\linewidth]{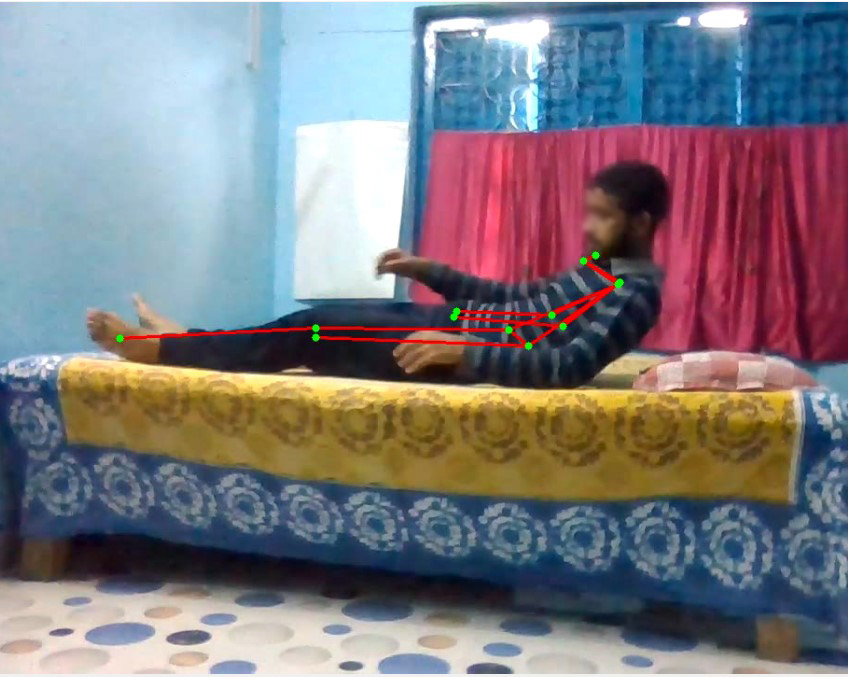}
 		\caption{ADL (04)}
 		\label{F:GMDCSA_1}
 	\end{subfigure}\hfil 
 	\begin{subfigure}{0.3\textwidth}
 		\includegraphics[width=\linewidth]{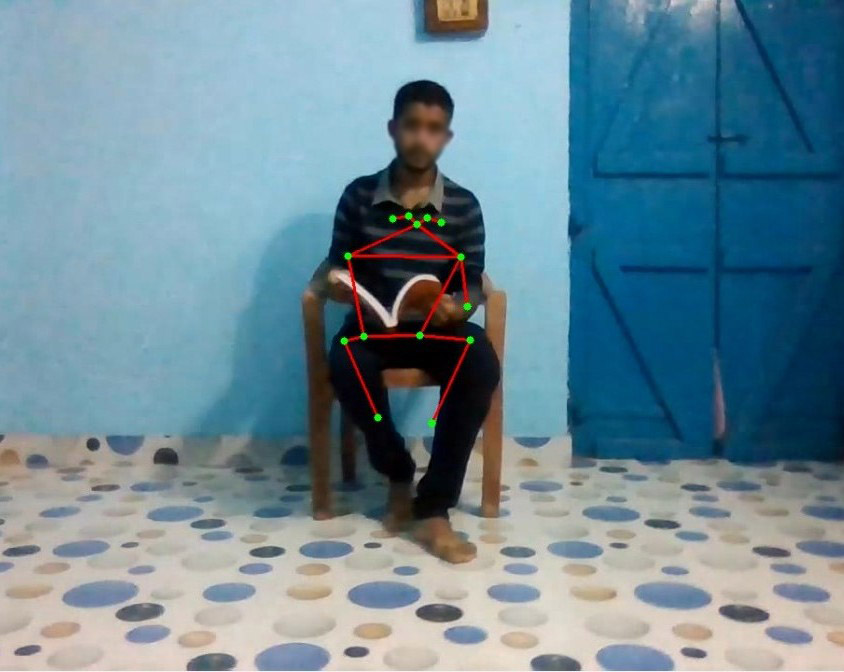}
 		\caption{ADL (07)}
 		\label{F:GMDCSA_2}
 	\end{subfigure}\hfil 
 	\begin{subfigure}{0.3\textwidth}
 		\includegraphics[width=\linewidth]{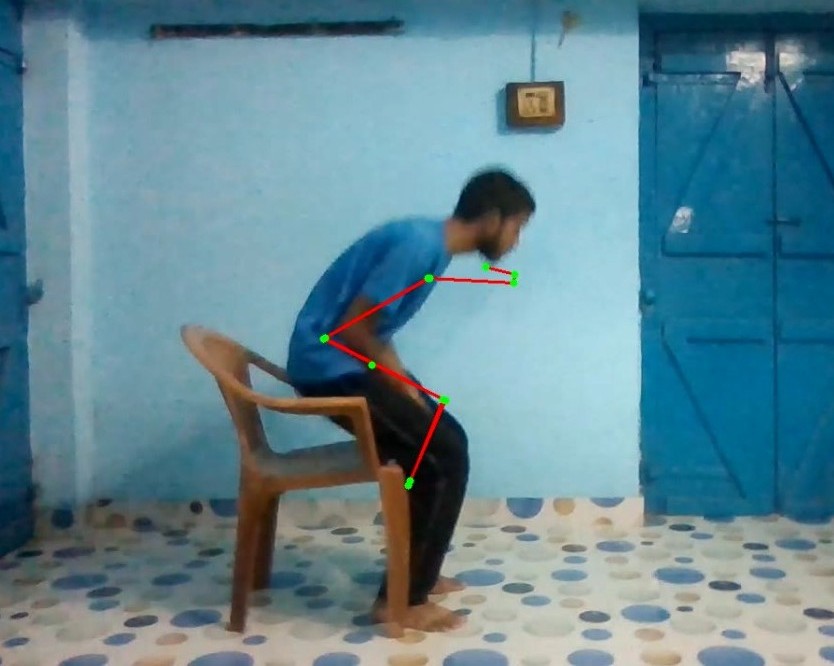}
 		\caption{ADL (14)}
 		\label{F:GMDCSA_3}
 	\end{subfigure}
 	
 	\medskip
 	\begin{subfigure}{0.3\textwidth}
 		\includegraphics[width=\linewidth]{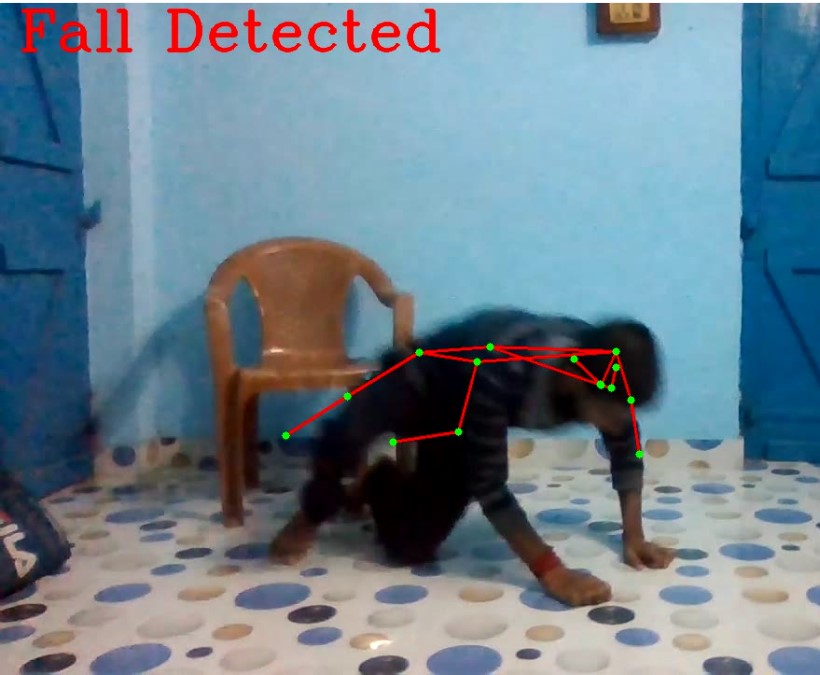}
 		\caption{Fall (03)}
 		\label{F:GMDCSA_4}
 	\end{subfigure}\hfil 
 	\begin{subfigure}{0.3\textwidth}
 		\includegraphics[width=\linewidth]{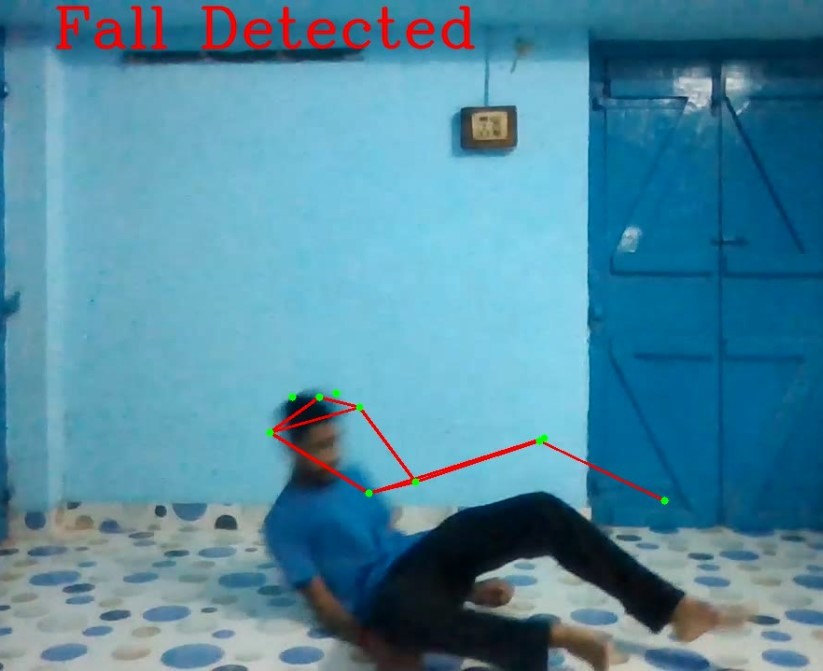}
 		\caption{Fall (12)}
 		\label{F:GMDCSA_5}
 	\end{subfigure}\hfil 
 	\begin{subfigure}{0.3\textwidth}
 		\includegraphics[width=\linewidth]{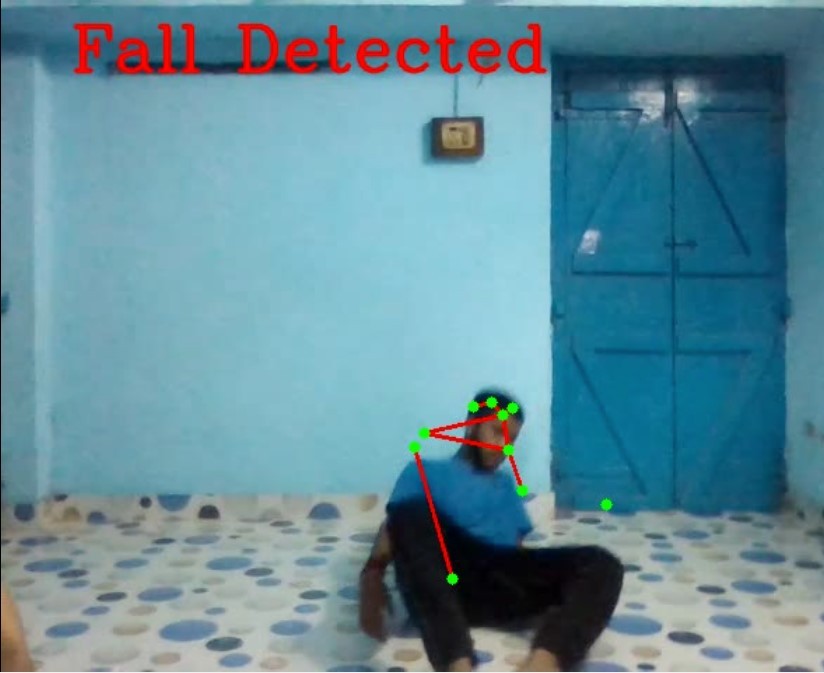}
 		\caption{Fall (14)}
 		\label{F:GMDCSA_6}
 	\end{subfigure}
 	\caption{Sample outputs using the GMDCSA dataset}
 	\label{F_GMDCSA}
 \end{figure}

 \begin{figure}[htb]
 	\centering 
 	\begin{subfigure}{0.3\textwidth}
 		\includegraphics[width=\linewidth]{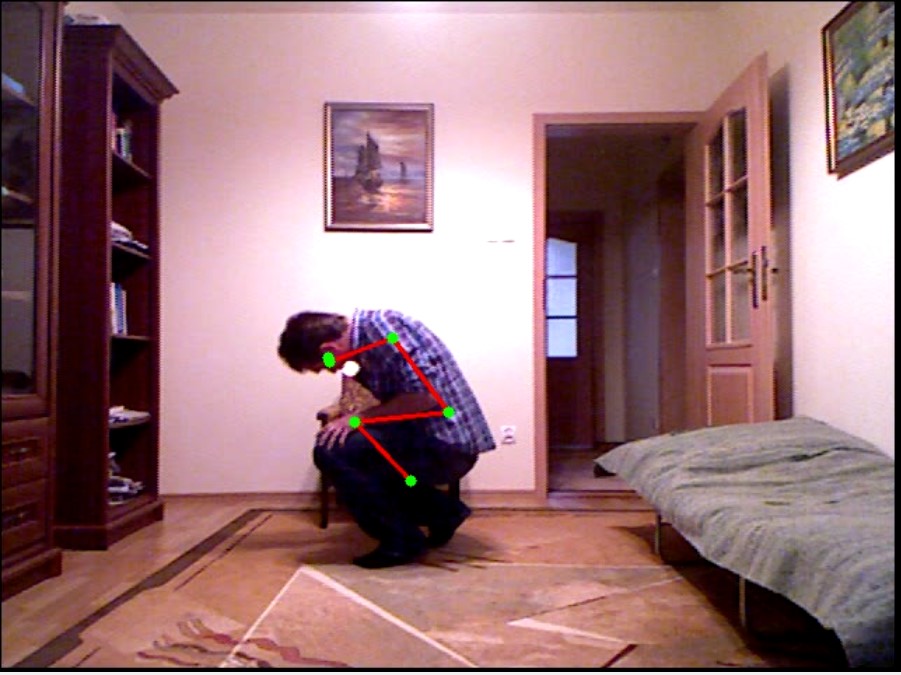}
 		\caption{ADL (01)}
 		\label{F_URFD_1}
 	\end{subfigure}\hfil 
 	\begin{subfigure}{0.3\textwidth}
 		\includegraphics[width=\linewidth]{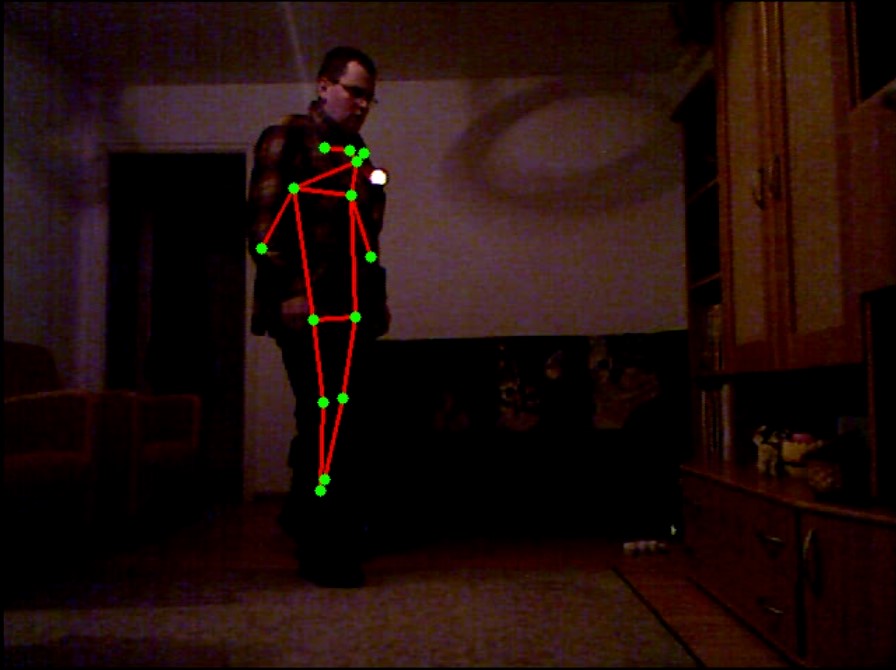}
 		\caption{ADL (12)}
 		\label{F_URFD_2}
 	\end{subfigure}\hfil 
 	\begin{subfigure}{0.3\textwidth}
 		\includegraphics[width=\linewidth]{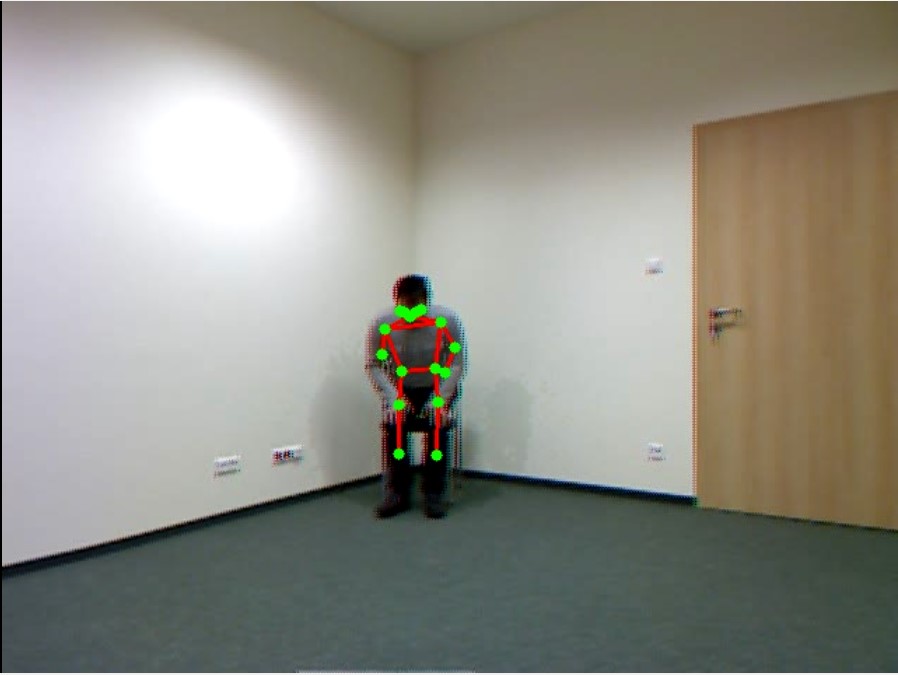}
 		\caption{ADL (25)}
 		\label{F_URFD_3}
 	\end{subfigure}
 	\medskip
 	
 	\begin{subfigure}{0.3\textwidth}
 		\includegraphics[width=\linewidth]{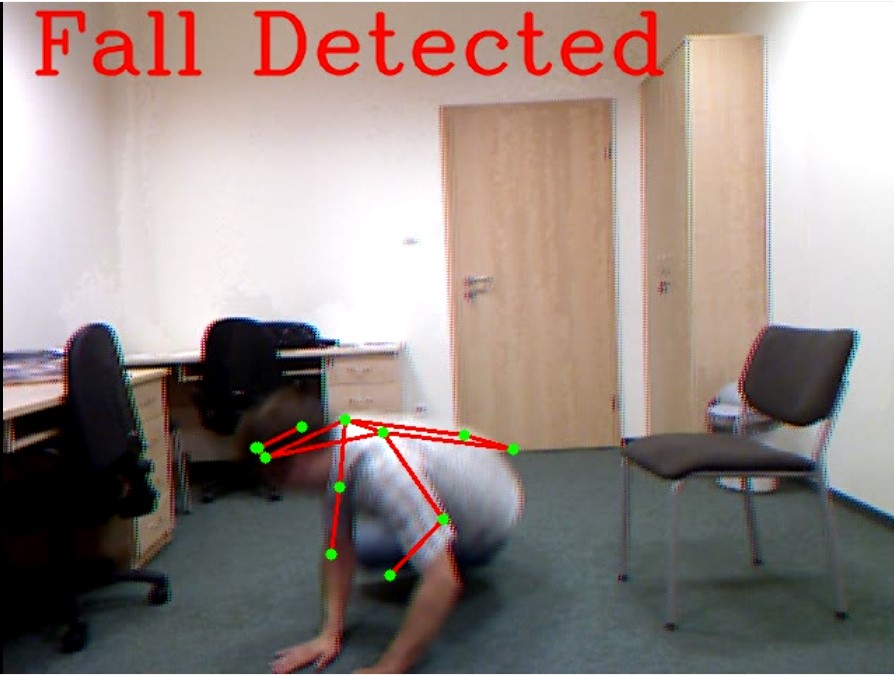}
 		\caption{Fall (01)}
 		\label{F_URFD_4}
 	\end{subfigure}\hfil 
 	\begin{subfigure}{0.3\textwidth}
 		\includegraphics[width=\linewidth]{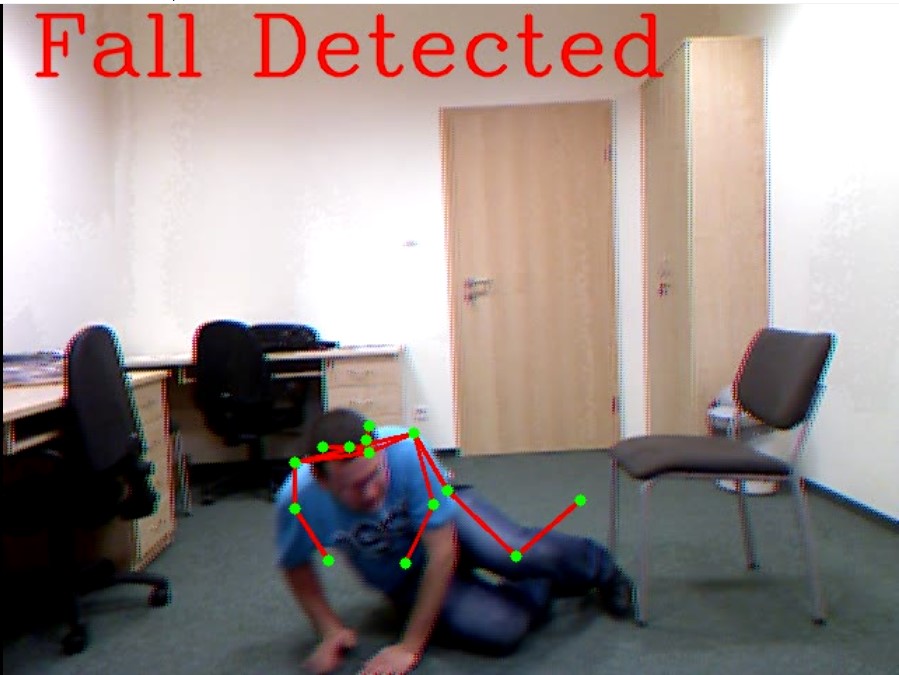}
 		\caption{Fall (08)}
 		\label{F_URFD_5}
 	\end{subfigure}\hfil 
 	\begin{subfigure}{0.3\textwidth}
 		\includegraphics[width=\linewidth]{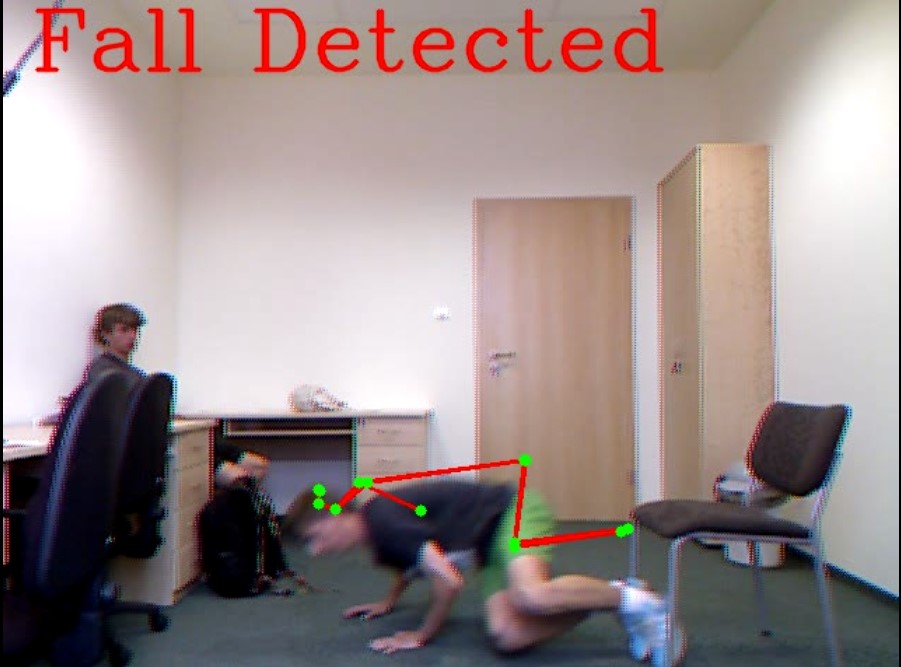}
 		\caption{Fall (14)}
 		\label{F_URFD_6}
 	\end{subfigure}
 	\caption{Sample outputs using the URFD dataset}
 	\label{F_URFD}
 \end{figure}
 Some of the sample outputs of this experiment using the GMDCSA and URFD datasets are shown in Figure \ref{F_GMDCSA} and \ref{F_URFD} respectively. The captions of the subfigure tell whether the frames are from the ADL sequence or the fall sequence. The number in the brackets of the caption is the file name of the video of the corresponding dataset. Figure \ref{F_GMDCSA} shows three sample ADL frames and three fall frames from the GMDCSA dataset video sequences. Similarly, Figure \ref{F_URFD} shows three sample ADL frames and three fall frames from the GMDCSA dataset video sequences.

\section{Conclusion and Future Scope}
\label{S_Conclusion}
In this paper, we proposed a lightweight and fast human fall detection system using `Movenet Thunder' pose estimation. Our proposed system is very fast and requires very low computing power. It can run easily in real-time on any low-computing device like mobile, laptop, desktop, etc. All computation is done locally, so it also preserves the privacy of the subject. The metrics are also good enough considering the low computing requirement of the system. The proposed technique gave good results using the GMDCSA dataset. The sensitivity values are good for both datasets. The Movenet pose estimation model is a fast and lightweight model, but its accuracy is moderate. Also, our system can not work for more than one subject at the same time. In the future, we are thinking to improve our system so that it can work in multi-person \cite{kocabas2018multiposenet} environments with high accuracy while maintaining the low computing requirement.

\bibliographystyle{elsarticle-num}
\scriptsize{\bibliography{RTHFD}}

\end{document}